\documentclass[conference]{IEEEtran}
\IEEEoverridecommandlockouts
\usepackage{cite}
\usepackage{amsmath,amssymb,amsfonts}
\usepackage{algorithmic}
\usepackage{graphicx}
\usepackage{textcomp}
\usepackage{xcolor}
\usepackage{subcaption}
\def\BibTeX{{\rm B\kern-.05em{\sc i\kern-.025em b}\kern-.08em
    T\kern-.1667em\lower.7ex\hbox{E}\kern-.125emX}}
\begin{document}

\title{ThinkDrive: Chain-of-Thought Guided Progressive Reinforcement Learning Fine-Tuning for Autonomous Driving}

\author{Chang Zhao, Zheming Yang, Yunqing Hu, Qi Guo, Zijian Wang, Pengcheng Li and Wen Ji}

\maketitle

\begin{abstract}
With the rapid advancement of large language models (LLMs) technologies, their application in the domain of autonomous driving has become increasingly widespread. However, existing methods suffer from unstructured reasoning, poor generalization, and misalignment with human driving intent. While Chain-of-Thought (CoT) reasoning enhances decision transparency, conventional supervised fine-tuning (SFT) fails to fully exploit its potential, and reinforcement learning (RL) approaches face instability and suboptimal reasoning depth. We propose ThinkDrive, a CoT guided progressive RL fine-tuning framework for autonomous driving that synergizes explicit reasoning with difficulty-aware adaptive policy optimization. Our method employs a two-stage training strategy. First, we perform SFT using CoT explanations. Then, we apply progressive RL with a difficulty-aware adaptive policy optimizer that dynamically adjusts learning intensity based on sample complexity. We evaluate our approach on a public dataset. The results show that ThinkDrive outperforms strong RL baselines by 1.45\%, 1.95\%, and 1.01\% on exam, easy-exam, and accuracy, respectively. Moreover, a 2B-parameter model trained with our method surpasses the much larger GPT-4o by 3.28\% on the exam metric.


\end{abstract}

\begin{IEEEkeywords}
Autonomous driving, LLMs, Chain-of-Thought, Reinforcement learning, Supervised fine-tuning
\end{IEEEkeywords}

\section{Introduction}
\label{sec:intro}

Recent advances in large language models (LLMs) have significantly enhanced performance in core capabilities, including perception, reasoning, and decision-making, establishing these models as fundamental components in autonomous driving research \cite{fu2024drive} \cite{petrovic2024llm}. They exhibit considerable promise in complex scene understanding, behavioral prediction, and trajectory planning. However, despite their strong performance on general tasks, persistent challenges remain in specialized autonomous driving scenarios \cite{Wei_2024_CVPR-XXX}. Current models often produce unstructured reasoning in their outputs, generalize poorly across diverse driving conditions, and fail to align with human driving intent. These limitations highlight the critical necessity for domain-specific model refinement tailored to autonomous driving systems.

\begin{figure}[tbp]
\centering
\includegraphics[width=\columnwidth]{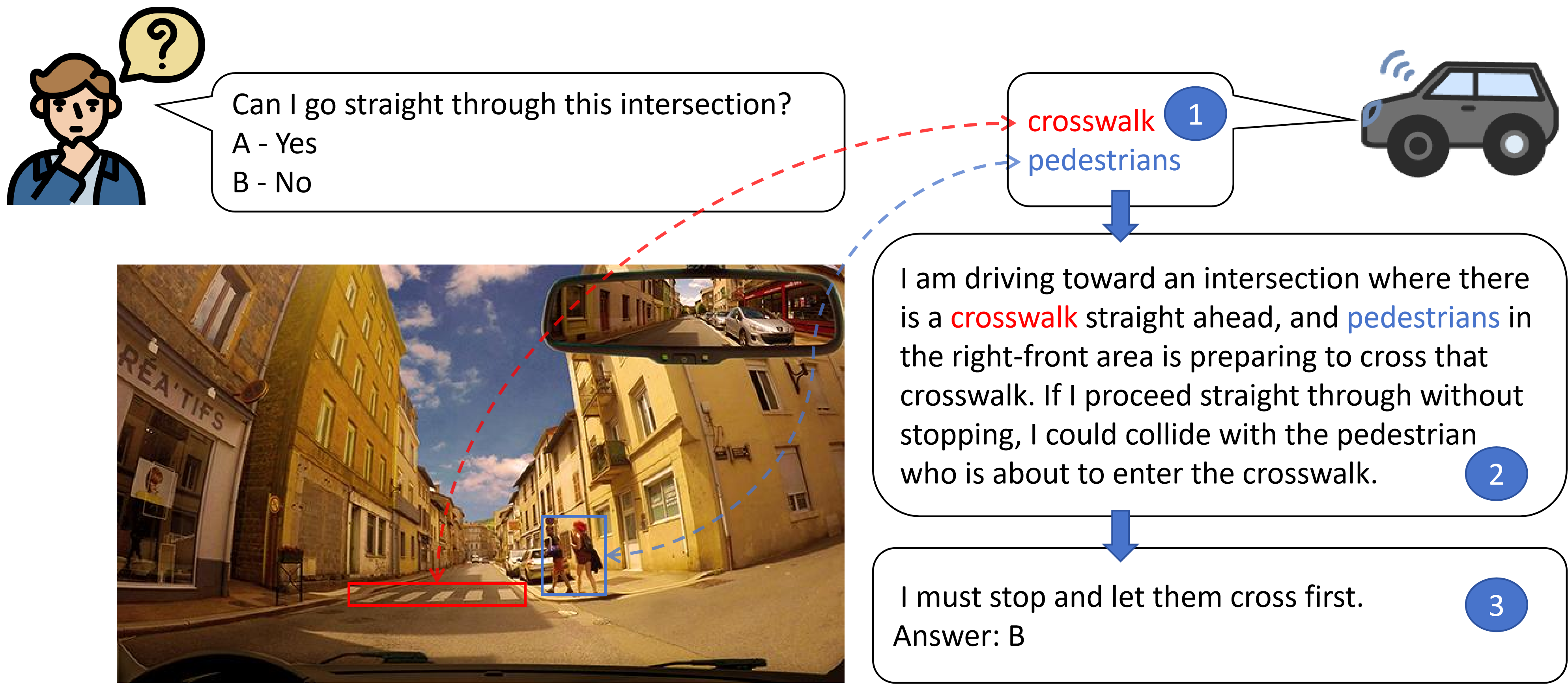}
\caption{Example of CoT reasoning applied in autonomous driving. The model first extracts entities from the image that are relevant to the question, then performs reasoning over these entities to derive the final answer.}
\label{fig:driving_task}
\end{figure}

Chain-of-Thought (CoT) is widely adopted in recent years to enhance decision transparency and logical rigor in autonomous driving systems \cite{wei2022chain}. Structured reasoning steps enable models to better comprehend causal relationships in complex traffic scenarios, thereby generating safer and more interpretable driving decisions, as illustrated in Fig.~\ref{fig:driving_task}. However, conventional supervised fine-tuning (SFT) methods often focus narrowly on aligning output formats when using CoT data. They fail to fully leverage the model’s intrinsic reasoning capabilities. As a result, these models show limited reasoning depth and poor generalization when facing dynamic, open-world driving environments. To overcome these limitations, researchers begin exploring RL fine-tuning approaches \cite{wang2022deep}. However, current RL methods still face significant practical challenges, such as unstable training. Moreover, the lack of progressive guidance during learning prevents models from fully realizing their reasoning potential in complex scenarios. This ultimately limits further performance improvements.

To address these challenges, we propose ThinkDrive, a CoT guided progressive RL fine-tuning framework for autonomous driving. Specifically, the framework first aligns the model with human-like driving rationales through CoT-based SFT. It then progressively improves reasoning quality and policy robustness via RL. During this stage, the optimization intensity is adaptively adjusted based on the difficulty of each training sample. The main contributions of this paper are as follows:

\begin{figure*}[t]
\centering
\includegraphics[width=\textwidth]{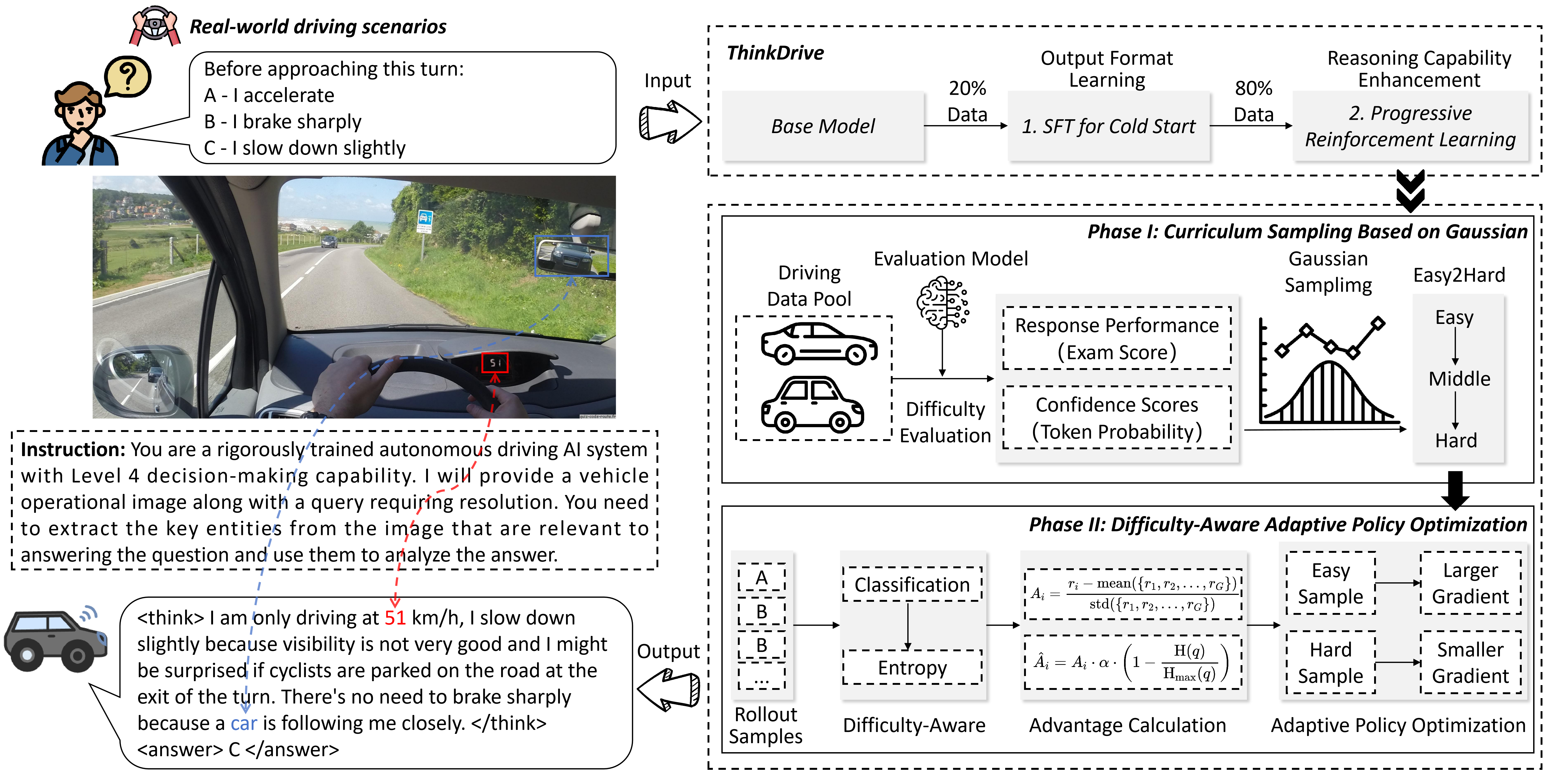}
\caption{Overview of the ThinkDrive framework for autonomous driving. The input consists of a driving scene image, the current query to be addressed, and task instructions. The model's reasoning capability is enhanced through a two-stage training strategy, where the most critical component is the progressive RL approach consisting of Gaussian-based curriculum sampling and the difficulty-aware adaptive policy optimization mechanism.}
\label{fig:framework}
\end{figure*}

\begin{itemize}
  \item We propose ThinkDrive, a CoT guided progressive RL fine-tuning framework for autonomous driving, which enhances the reasoning capabilities of multimodal LLMs in autonomous driving through CoT prompting and a two-stage training strategy. 

  \item We design a difficulty-aware adaptive policy optimization algorithm that dynamically adjusts the magnitude of policy optimization based on sample difficulty. This approach enables more stable and efficient RL training, thereby more effectively eliciting the model's deep reasoning capabilities.

  \item Experiments demonstrate that our method outperforms current state-of-the-art RL algorithms by 1.45\%, 1.95\%, and 1.01\% on the exam, easy-exam, and accuracy metrics, respectively. Moreover, a 2B-parameter model trained with our method surpasses the significantly larger GPT-4o by 3.28\% on the exam metric.
\end{itemize}

\section{Related Works}

LLMs exhibit strong capabilities in semantic understanding \cite{ADA-XXXX}, inductive reasoning, and knowledge generalization, which have introduced a new paradigm for autonomous driving research \cite{naveed2025comprehensive}. This paradigm employed LLMs to process sensor information and integrated common-sense priors such as traffic rules to support driving decision-making \cite{fu2024drive}. Yang et al. \cite{yang2024driving} further demonstrated that LLMs can align autonomous agents with human driving styles. To enhance environmental perception, multimodal LLMs incorporated inputs from cameras, LiDAR, and other sensors \cite{yin2024survey,lu2025omnitester,cui2024survey}. As driving tasks grow more complex, CoT reasoning has been introduced to emulate human-like sequential cognition and improve decision-making. For example, DriveLM \cite{sima2024drivelm} decomposed driving tasks into subtasks and guided stepwise inference to generate decisions, showing strong potential in autonomous driving systems \cite{10980428}. Nevertheless, reliably training models to produce correct CoT and consistent decisions remains a significant challenge.

The success of DeepSeek-R1 \cite{guo2025deepseek} demonstrated that RL effectively enhances model reasoning capabilities, making it particularly suitable for CoT–dependent scenarios. Among recent approaches, GRPO \cite{shao2024deepseekmath} and its variants have attracted significant attention. GRPO adopted an intra-group relative ranking mechanism that guides policy optimization toward advantage maximization. However, it inherently suffers from issues such as entropy collapse and length bias. To address these issues, several extensions have been proposed. DAPO \cite{yu2025dapo} alleviated entropy collapse by expanding exploration over low-probability tokens through a clip-higher mechanism. GSPO \cite{zheng2025gspo} improved training efficiency using sequence-level importance sampling ratios. GMPO \cite{zhao2025geometric} reduced sensitivity to outliers by replacing arithmetic averaging with geometric mean aggregation. Despite these advances, existing methods still exhibit training instability and fail to fully activate the model’s reasoning potential.

\section{The Proposed ThinkDrive Framework}

We propose ThinkDrive, a CoT guided progressive RL fine-tuning framework for autonomous driving, as illustrated in Fig.~\ref{fig:framework}. The framework aims to enhance the decision-making capabilities of multimodal LLMs in complex driving scenarios by integrating CoT data with a two-stage training strategy. The first stage employs SFT on CoT data to initialize the model, yielding a cold-start model with preliminary structured reasoning capabilities. Building upon the cold-start model obtained in the first stage, the second stage incorporates curriculum learning \cite{narvekar2020curriculum} into the RL phase to enable progressive policy optimization. Specifically, curriculum sampling first categorizes the training data based on confidence scores and response performance, then employs a Gaussian function to smoothly interpolate between data of different difficulty levels. Furthermore, we refine the policy loss function of the GRPO algorithm to enhance training stability and improve overall performance.

\subsection{Progressive Reinforcement Learning}
Following the cold-start phase of SFT, the model can generate outputs that conform to the expected format. However, its deeper reasoning capabilities remain underutilized, thereby limiting decision-making performance. To overcome this bottleneck, the second stage incorporates RL for policy refinement to improve the model’s potential for reasoning in complex scenarios. Nevertheless, existing algorithms based on group-relative policy optimization face significant challenges during early training stages. During early training, the model lacks sufficient reasoning capability. When training batches contain a random mixture of driving scenarios with varying difficulty levels, it struggles to generate correct CoT reasoning and accurate outputs for complex samples. This typically introduces noise into policy optimization, resulting in training instability and even policy collapse.


To address this issue, we integrate curriculum learning with RL in the second stage to enable progressive RL fine-tuning. The first step involves partitioning the training data into subsets of varying difficulty levels. To achieve this, we perform SFT on the full dataset to train an evaluation model. Notably, we exclude CoT annotations from the training data. This is because reasoning models trained with CoT data tend to generate reasoning traces that heavily bias the final answer distribution. As a result, the predicted answers often exhibit near-deterministic token probabilities. This undermines the reliability of model confidence as a meaningful uncertainty measure.

Then, we use the evaluation model to perform inference on the dataset designated for the RL phase, obtaining both the model’s predictions and their associated confidence scores. Based on the confidence scores and prediction accuracy, we categorize the data into three difficulty levels: easy, medium, and hard. The following is the difficulty classification rule:

\begin{equation}
d_i=\begin{cases}\text{Easy,}&prediction_i=label_i\:\text{and}\:c_i\geq th_1.\\\text{Hard,}&prediction_i \neq label_i\:\text{and}\:c_i\geq th_2.\\\text{Medium,}&\text{otherwise}.\end{cases}
\end{equation}

Where $c_i$ denotes the confidence score of the model’s predicted answer, and ${th}_1$ and ${th}_2$ are user-defined confidence thresholds. If the model’s prediction is correct and its confidence score is high, the sample is classified as easy, indicating that the model can reliably produce the correct prediction for this instance. Conversely, samples for which the model makes an incorrect prediction yet assigns a high confidence score are categorized as hard. All remaining samples are labeled as medium.

Directly concatenating data with varying difficulty levels can cause abrupt transitions in task complexity at segment boundaries \cite{parashar2025curriculum}. This may still lead to training instability. To ensure a smoother progression, we introduce a Gaussian-weighted function to construct a difficulty scheduler. This scheduler prioritizes policy optimization on easy samples during the early stages of training. It then gradually shifts emphasis toward more challenging samples in a continuous and controlled manner. This design helps prevent abrupt policy shifts. The formulations of the scheduler are as follows:


\begin{equation}
\mathbf{S}_{\text{Gaussian}}(t,k) = \exp\left(-\frac{(x_{t}-\mu_{k})^{2}}{2\sigma^{2}}\right),
\end{equation}

\begin{equation}
x_{t} = \left(\frac{t}{T}\right)^{\beta}(K-1),
\end{equation}
the variance $\sigma$ controls the concentration of sampling distribution, while $\beta$ governs how rapidly the sampling position $x_t$ evolves over time.


After obtaining data whose difficulty varies with training steps, we apply RL to the curriculum-sampled data. This enables the model to learn to solve autonomous driving tasks of varying difficulty progressively. The objective function of the RL is formulated as follows, aiming to maximize the expected cumulative return induced by the policy.

\begin{equation}
J(\theta) = \mathbb{E}_{\tau \sim \pi_\theta} \left[ \sum_{t=0}^{\infty} \gamma^t R_{t+1} \right],
\end{equation}
where $\pi_\theta$ denotes the policy parameterized by $\theta$, $\tau$ represents a trajectory sampled from the policy, and $\gamma$ is the discount factor that balances immediate and long-term rewards. By integrating curriculum learning with reinforcement learning, this progressive RL strategy effectively constrains the policy optimization process within an adaptive difficulty range, thereby reducing the variance of gradient updates and improving overall training stability. Moreover, the gradual exposure to increasingly complex scenarios allows the model to refine its internal reasoning process and decision-making strategies in a structured manner.




\subsection{Difficulty-Aware Adaptive Policy Optimization}
GRPO \cite{shao2024deepseekmath} inherently suffers from training instability, limiting its ability to fully unlock model capabilities. This issue is particularly pronounced in autonomous driving scenarios, where decision-making difficulty varies substantially across diverse weather conditions and road situations. This variation further exacerbates training instability. Although integrating curriculum learning helps reduce intra-batch difficulty disparities to some extent, it lacks fine-grained adaptability and fails to account for the dynamic evolution of sample difficulty during training.

To further stabilize training and enhance RL fine-tuning efficacy, we propose a difficulty-aware adaptive policy optimization method. Our approach is motivated by the observation that complex samples induce high model uncertainty, leading to divergent outcomes across multiple rollouts, whereas simple samples consistently yield identical decisions. To quantify sample difficulty, we compute the entropy of multi-rollout outputs. For each sample, we first cluster rollout trajectories according to their decision outcomes. We then sum the probabilities of trajectories within each cluster to obtain cluster-level probabilities. To mitigate estimation bias caused by a limited number of rollouts, we normalize these cluster probabilities. Finally, we compute the entropy of the resulting normalized distribution. The formulation is presented as follows:



\begin{equation}
\mathrm{H}(q) = -\sum_{c \in \mathcal{C}} 
\left( 
\frac{\sum_{o_i \in c} p(o_i \mid q)}{\sum_{j=1}^{G} p(o_j \mid q)} 
\log 
\left( 
\frac{\sum_{o_i \in c} p(o_i \mid q)}{\sum_{j=1}^{G} p(o_j \mid q)} 
\right) 
\right).
\end{equation}


Where $c$ denotes decision categories. A high entropy indicates a high-difficulty sample, for which the magnitude of policy optimization should be reduced to mitigate training instability. Conversely, low-entropy samples reflect high model confidence, allowing for a larger policy optimization magnitude. To enable sample-level control over policy optimization, we incorporate the computed entropy into the advantage estimation. This indirectly modulating the optimization magnitude. The specific formulation is as follows:

\begin{equation}
\hat{A}_i=\frac{R_i - \mathrm{mean}(\{R_i\}_{i=1}^G)}{\mathrm{std}(\{R_i\}_{i=1}^G)}\cdot \alpha\cdot \big(1-\mathrm{H}(q)/\mathrm{H}_{\max}\big).
\end{equation}


$H_{\max}$ denotes the maximum possible entropy, which is given by $-\log(G)$, where $G$ is the number of rollouts. Dividing by $H_{\max}$ normalizes the entropy ratio $\mathrm{H}(q)/\mathrm{H}_{\max}$ to the interval $[0, 1]$. The hyperparameter $\alpha$ is a user-defined scaling coefficient that controls the extent to which the policy optimization magnitude is amplified or attenuated.




Furthermore, to avoid the constraint imposed by the reference model on the exploration space, we omit the reference model entirely. To enhance exploration of low-probability tokens and prevent premature entropy collapse, we adopt the clip-higher mechanism from DAPO \cite{yu2025dapo}. Additionally, to address the issue of invalid samples caused by model outputs consisting entirely of 0 or 1, we introduce dynamic sampling. Considering the adverse impact of noisy samples, we further improve robustness by replacing the arithmetic averaging used in \cite{shao2024deepseekmath} with the geometric mean proposed in GMPO \cite{zhao2025geometric}, which contributes to more stable training dynamics. The final objective function of our method is as follows:



\begin{equation}
\begin{aligned}
\mathcal{J}_{\mathrm{ours}}(\pi_\theta) &=\mathbb{E}_{(q,a)\thicksim\mathcal{D},\{o_i\}_{i=1}^G\thicksim\pi_{\theta_{\mathrm{old}}}(\cdot \mid q)}\\&\frac1G\sum_{i=1}^G\left\{\prod_{t=1}^{|o_i|}\left(\min\left[r_{i,t}(\theta),clip_{\text{higher}}(\theta)\right]\right)\right\}^{\frac1{|o_i|}}\hat{A}_i \\ 
\mathrm{s.t.}\quad & \quad0<\left|\{o_i\mid\text{is equivalent}(a,o_i)\}\right|<G,
\end{aligned}
\end{equation}
where $r_{i,t}(\theta)$ is the importance sampling weight, which is defined as $\frac{\pi_\theta(o_{i,t}|q,o_{i,<t})}{\pi_{\theta_{\mathrm{old}}}(o_{i,t}|q,o_{i,<t})}$. And $clip_{\text{higher}}(\theta)$ employs an asymmetric clipping strategy to mitigate the entropy collapse issue. The specific formulation is as follows:
\begin{equation}
clip_{\text{higher}}(\theta) = \mathrm{clip}\left(r_{i,t}(\theta),1-\epsilon_{low},1+\epsilon_{high}\right).
\end{equation}


The proposed difficulty-aware adaptive policy optimization method can improve training stability during the RL phase. It also effectively elicits the model’s deep reasoning capabilities, leading to enhanced decision-making accuracy in autonomous driving.


\section{Experiments}

\subsection{Experimental Setup}
We employ DrivingVQA \cite{corbiere2025drivingvqa}, an autonomous driving question-answering dataset comprising real-world driving scenarios, where all questions are presented in multiple-choice format, and each answer is accompanied by a CoT explanation. We use the Qwen3-VL-2B model as the base model, and all training experiments are conducted using this base model as the initial model. Given that the dataset contains single-choice, multiple-choice, and multi-question samples, the SFT phase uses a subset constructed by randomly selecting 20\% of samples from each question type and concatenating them. The remaining 80\% of the data is reserved for the second-stage progressive RL. In the second stage, the curriculum sampling strategy is defined with confidence thresholds $th_1 = 0.6$ and $th_2 = 0.4$, the Gaussian weighting function parameterized by $\beta = 0.5$ and $\sigma = 0.5$, and the weight $\alpha$ for computing the advantage is set to $1.5$. The reward functions used in all RL experiments are easy-exam reward and format reward. The easy-exam reward is computed identically to the easy-exam metric, as it provides finer-grained feedback compared to the strict exam metric, thereby facilitating more effective policy optimization. Experiments are conducted on a server equipped with two NVIDIA A100 40GB GPUs. To enable efficient training, we employed the LLaMA-Factory framework for SFT and the verl framework for RL training. The training environment utilized CUDA 12.2, Python 3.10, and PyTorch 2.6.0. For model evaluation, we adopt three metrics: exam (requiring all correct options to be selected for each question), easy-exam (granting partial credit for partially correct selections in multiple-choice questions), and accuracy.

\begin{figure}[htbp] 
    \centering
    \begin{subfigure}[b]{0.50\columnwidth} 
        \centering
        \includegraphics[width=\linewidth]{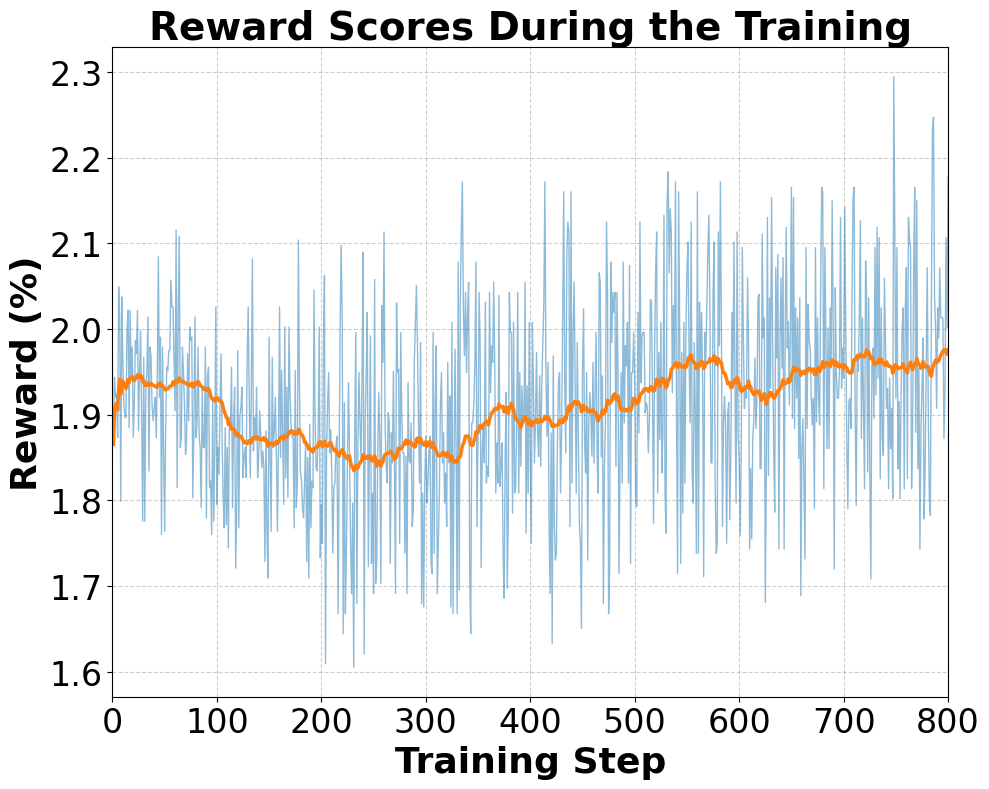}
        \caption{}
        \label{fig:sub1}
    \end{subfigure}
    \hfill
    \begin{subfigure}[b]{0.50\columnwidth}
        \centering
        \includegraphics[width=\linewidth]{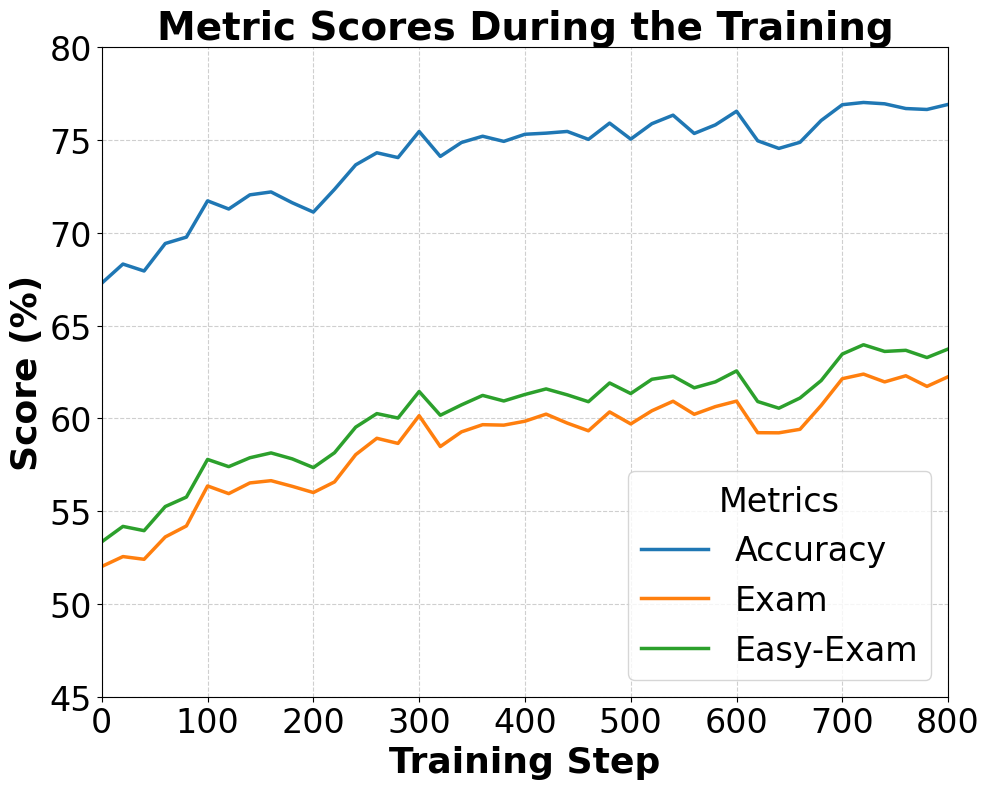}
        \caption{}
        \label{fig:sub2}
    \end{subfigure}
    
    \caption{Training dynamics of key metrics with ThinkDrive. (a) Reward score (smoothed using a rolling average over 50 steps to reduce noise and emphasize the overall trend); (b) accuracy, exam, and easy-exam scores.}
    \label{fig:train_self}
\end{figure}

\subsection{Evaluation Results}
\subsubsection{Convergence Analysis}
We train the Qwen3-VL-2B model using the ThinkDrive framework in a two-stage training strategy and record the evolution of key metrics during training, as shown in Fig.~\ref{fig:train_self}. The training reward initially decreases before gradually increasing and stabilizing. At the beginning, the high proportion of easy samples leads to a high initial reward. However, as the difficulty of sampled data increases, the model’s limited capability causes a temporary drop in reward. Subsequently, as the model’s reasoning ability improves, the reward steadily rises and eventually remains high even when exposed to difficult samples, demonstrating the effectiveness of the training process. Moreover, the exam, easy-exam, and accuracy metrics all exhibit steady improvement and ultimately converge, indicating that the model’s reasoning capability progressively strengthens throughout training.

\begin{figure}[th] 
    \centering
    \begin{subfigure}[b]{0.5\columnwidth} 
        \centering
        \includegraphics[width=\linewidth]{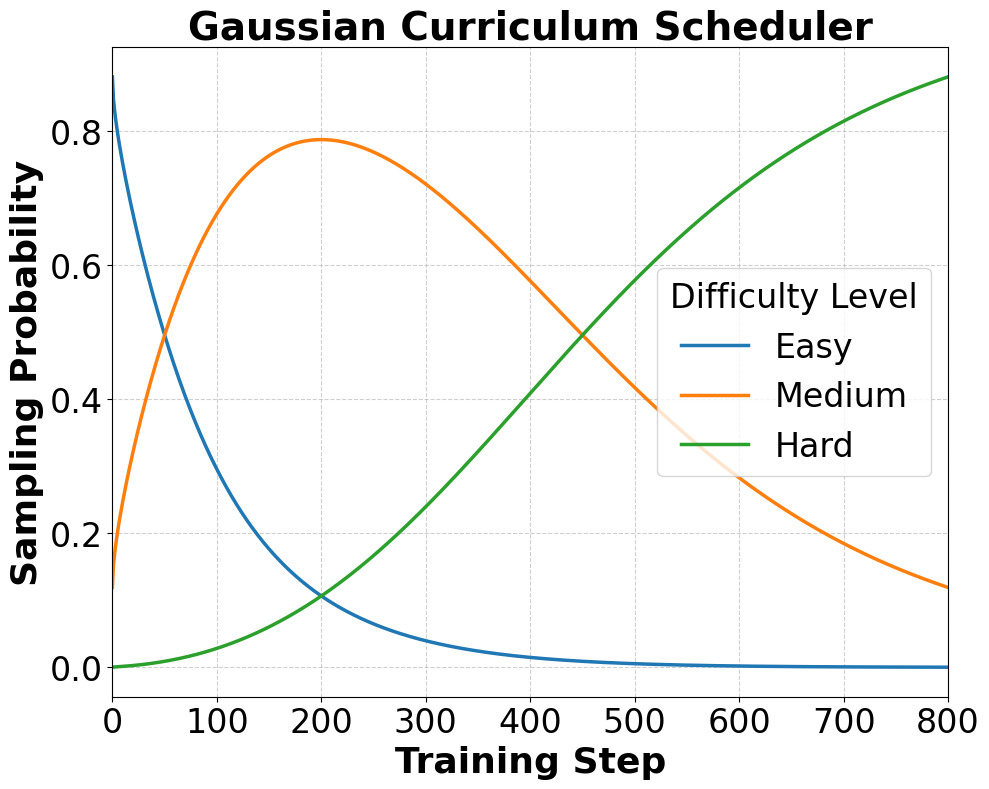} 
        \caption{} 
        \label{fig:left}
    \end{subfigure}
    \hfill 
    \begin{subfigure}[b]{0.5\columnwidth}
        \centering
        \includegraphics[width=\linewidth]{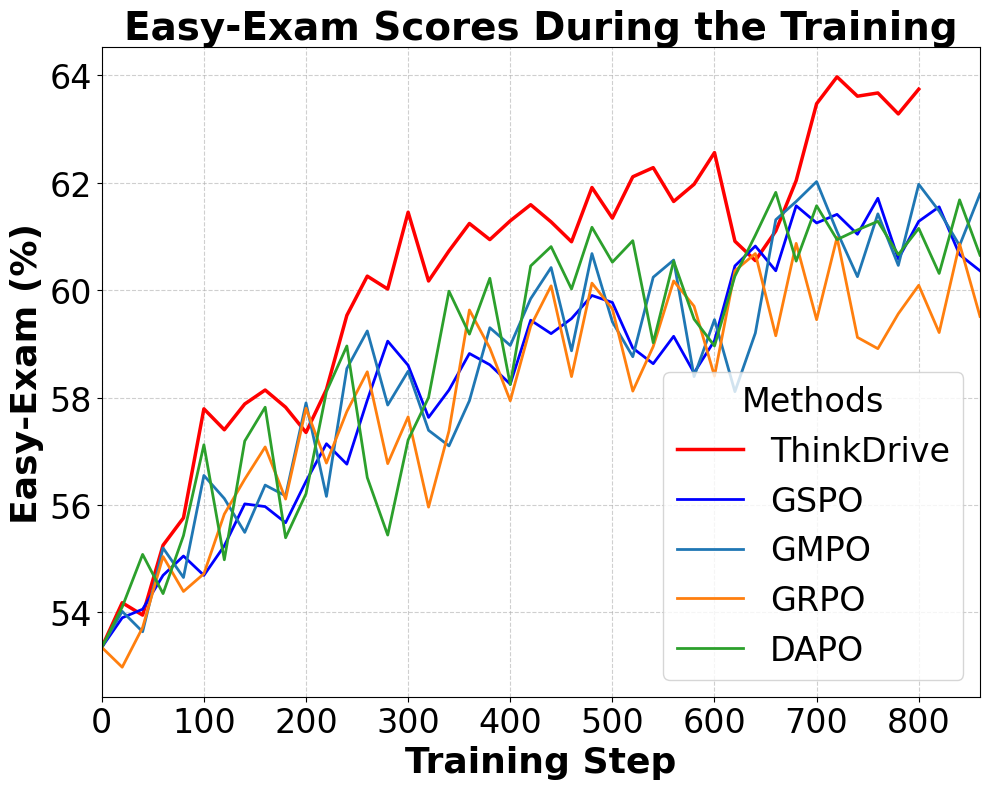}
        \caption{} 
        \label{fig:right}
    \end{subfigure}
    
    \caption{Variation of data difficulty and easy-exam metric with training steps. (a) Proportion of samples from each difficulty level at different training steps. (b) Comparison of easy-exam metric across different methods during training.}
    \label{fig:easy_exam}
\end{figure}

Fig.~\ref{fig:easy_exam} shows the proportions of data from different difficulty levels used at each training step and the comparison of the easy-exam metric across different methods during training. Our method exhibits significantly reduced training instability and converges more efficiently than other RL approaches, demonstrating both enhanced training stability and improved sample efficiency. These observations indicate that the performance gains of ThinkDrive arise not from stronger optimization alone, but from the synergistic integration of difficulty-aware adaptive policy optimization and curriculum learning. By dynamically aligning learning intensity with sample complexity, the proposed framework suppresses oscillatory training behaviors and enables more efficient utilization of training data.

\subsubsection{End-to-End Performance}

We compare our method to SFT and several state-of-the-art RL algorithms. To isolate the effect of the optimization strategy, each RL training is initialized from an identical cold-start model. The experimental results are summarized in Table~\ref{tab:rl_comparison}. Our method consistently outperforms all baselines across three evaluation metrics: exam, easy-exam, and accuracy. Specifically, compared to SFT, our approach achieves absolute improvements of 4.36\%, 5.29\%, and 2.44\% on exam, easy-exam, and accuracy, respectively. Furthermore, it surpasses GMPO, the strongest RL baseline, by 1.45\%, 1.95\%, and 1.01\% on the same metrics. These results show that ThinkDrive, leveraging its progressive RL approach, significantly enhances the model’s reasoning capabilities compared to both SFT and other RL methods. By adaptively adjusting learning intensity according to sample complexity, ThinkDrive enables deeper reasoning refinement on challenging scenarios. It can produce more robust and generalizable decisions, achieving balanced improvements in both reasoning correctness and overall decision accuracy.

\begin{table}[htbp]
\centering
\caption{Comparison of ThinkDrive with other training methods. The table highlights the superior performance of ThinkDrive compared to other state-of-the-art methods such as GRPO, DAPO, GMPO and GSPO.}
\label{tab:rl_comparison}
\begin{tabular}{lcccc}
\hline
Method & Exam(\%) & Easy-Exam(\%) & Accuracy(\%)\\
\hline
SFT & 58.02 & 58.68 & 74.58 \\
GRPO\cite{shao2024deepseekmath} & 60.27 & 60.98 & 75.09 \\
GSPO\cite{zheng2025gspo} & 60.41 & 61.71 & 75.38 \\
DAPO\cite{yu2025dapo} & 60.55 & 61.82 & 75.82 \\
GMPO\cite{zhao2025geometric} & 60.93 & 62.02 & 76.01\\
\hline
\textbf{ThinkDrive} & \textbf{62.38} & \textbf{63.97} & \textbf{77.02}\\
\hline
\end{tabular}
\end{table}

\begin{figure}[htbp]
\centering
\includegraphics[width=\columnwidth]{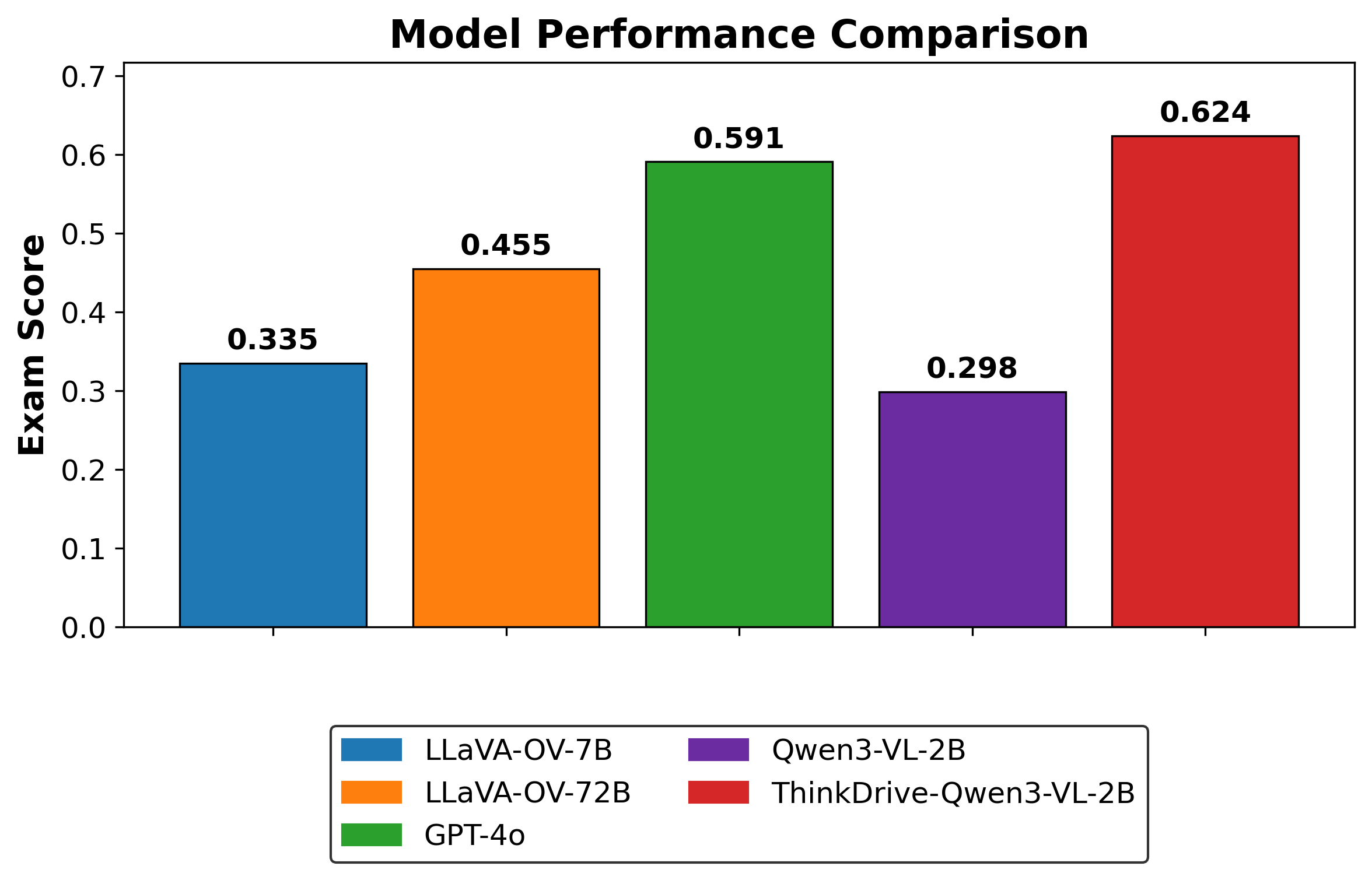}
\caption{Comparison of exam metrics for open-source models.}
\label{fig:open_model}
\end{figure}

Moreover, we compare our approach to several open-source models, with results reported in Fig.~\ref{fig:open_model}. After training with our method, the 2B-parameter model outperforms the much larger GPT-4o \cite{islam2025gpt} by 3.28\% on the exam metric. This result highlights that performance in autonomous driving reasoning is not solely determined by model capacity but critically depends on how effectively reasoning structures are learned and optimized. While large-scale models benefit from extensive pretraining, they may exhibit misaligned or overly generic reasoning patterns when directly applied to domain-specific driving tasks. In contrast, ThinkDrive explicitly aligns the model’s decision process with structured CoT reasoning and progressively refines it via difficulty-aware RL, enabling smaller models to develop more task-relevant and reliable reasoning behaviors.

\subsection{Ablation Studies}
To empirically assess the contribution of each component in the proposed ThinkDrive framework, we perform an ablation study by decoupling it into three modules: SFT, difficulty-aware adaptive policy optimization, and Gaussian-based curriculum learning. These components are incrementally integrated, and the results are summarized in Table~\ref{tab:Ablation}. It can be found that the difficulty-aware adaptive policy optimization algorithm alone improves the post–cold-start model by 9.56\%, 9.75\%, and 9.26\% on the exam, easy-exam, and accuracy metrics, respectively. Compared to GMPO, the best-performing method in prior experiments, it yields further gains of 0.65\%, 1.08\%, and 0.54\% on the same metrics. Incorporating the Gaussian-based curriculum learning module provides additional improvements of 0.80\%, 0.87\%, and 0.47\%. This indicates that curriculum learning plays a complementary role by stabilizing the learning process and facilitating smoother transitions across difficulty levels. The ablation study clearly validates that the difficulty-aware mechanism serves as the core driver of performance improvement, while the auxiliary role of curriculum learning further refines the model’s reasoning capability. Collectively, these results demonstrate the necessity of each component in the proposed ThinkDrive framework.



\begin{table}[htbp]
\centering
\caption{Decomposition of ThinkDrive and contribution of each component.}
\label{tab:Ablation}
\begin{tabular}{lccc}
\hline
Model & Exam(\%) & Easy-Exam(\%) & Accuracy(\%) \\
\hline
Qwen3-VL-2B & 29.84 & 31.22 & 58.46 \\
+ SFT (Cold Start) & 52.02 & 53.35 & 67.29 \\
+ Difficulty-Aware RL & 61.58 & 63.1 & 76.55 \\
+ Curriculum Learning & 62.38 & 63.97 & 77.02 \\
\hline
\end{tabular}
\end{table}

\section{Conclusion}
This paper presents ThinkDrive, a CoT guided progressive reinforcement learning framework for enhancing the reasoning capability of multimodal LLMs in autonomous driving. By combining CoT-based supervised fine-tuning with a difficulty-aware adaptive policy optimization mechanism and a Gaussian-based curriculum learning strategy, ThinkDrive enables structured and stable reasoning acquisition across driving scenarios of varying complexity. In contrast to conventional SFT and existing RL-based methods, the proposed framework effectively mitigates training instability while more fully eliciting the model’s latent reasoning potential. Extensive experimental results demonstrate that ThinkDrive consistently outperforms strong RL baselines across multiple evaluation metrics. These results validate the effectiveness of ThinkDrive as a robust and scalable method for CoT reasoning in autonomous driving.





\bibliographystyle{IEEEbib}
\bibliography{icme2026references}

\vspace{12pt}

\end{document}